\documentclass[conference]{IEEEtran}
\IEEEoverridecommandlockouts
\usepackage{cite}
\usepackage{amsmath,amssymb,amsfonts}
\usepackage{algorithmic}
\usepackage{graphicx}
\usepackage{textcomp}
\usepackage{xcolor}
\usepackage{float}
\def\BibTeX{{\rm B\kern-.05em{\sc i\kern-.025em b}\kern-.08em
    T\kern-.1667em\lower.7ex\hbox{E}\kern-.125emX}}
\begin{document}

\title{Collaborative Perception in Multi-Robot Systems: Case Studies in Household Cleaning and Warehouse Operations\\

%
}

\author{\IEEEauthorblockN{Bharath Rajiv Nair}
\IEEEauthorblockA{\textit{Dept. of Mechanical Engineering} \\
\textit{Columbia University}\\
New York, USA \\
bharathnair2637@gmail.com }

}

\maketitle

\begin{abstract}
This paper explores the paradigm of Collaborative Perception (CP), where multiple robots and sensors in the environment share and integrate sensor data to construct a comprehensive representation of the surroundings. By aggregating data from various sensors and utilizing advanced algorithms, the collaborative perception framework improves task efficiency, coverage, and safety. Two case studies are presented to showcase the benefits of collaborative perception in multi-robot systems. The first case study illustrates the benefits and advantages of using CP for the task of household cleaning with a team of cleaning robots. The second case study performs a comparative analysis of the performance of CP versus Standalone Perception (SP) for Autonomous Mobile Robots operating in a warehouse environment. The case studies validate the effectiveness of CP in enhancing multi-robot coordination, task completion, and overall system performance and its potential to impact operations in other applications as well. Future investigations will focus on optimizing the framework and validating its performance through empirical testing.
\end{abstract}


\section{Introduction}

In recent years, the field of robotics has seen significant progress, particularly in the context of multi-robot systems. Collaborative perception, where multiple robots and sensors in the environment share and integrate sensory data to create a unified understanding of their environment, has emerged as a crucial area of research. This collaborative approach to perception enhances the efficiency, accuracy, and safety of robotic operations, making it highly applicable across a variety of domains such as industrial automation, autonomous driving, household cleaning and environmental monitoring \cite{1_han2023collaborative, 2_cortes2017coordinated, 3_vicentini2021collaborative}.

The paradigm of collaborative perception enhances the capabilities of a robot to perceive it's environment by compensating for the limits of its individual sensing capabilities such as limited range and field of view. This in turn enables the robots to coordinate their actions more effectively and perform individual tasks with greater efficiency. This paper presents collaborative perception frameworks designed to facilitate multi-robot collaboration, focusing on two specific case studies: household cleaning using a team of cleaning robots and autonomous mobile robots (AMRs) operating in warehouse environments.

In the first case study, we explore a scenario where a team of cleaning robots, including a robotic vacuum cleaner, a staircase cleaning robot, and a trash picking robot, work together to clean a home environment. Overhead cameras are strategically placed to provide a global view of the environment. Using this collaborative perception setup, the robots can coordinate their actions, identify areas that require cleaning, and delegate tasks dynamically based on the changes in the environment and the capabilities of each robot.

The second case study examines AMRs in a warehouse setting, where individual robots may not be aware of all obstacles in its planned path due to occlusions and limitations in its sensing range. By utilizing overhead sensors, the framework allows these robots to receive real-time updates about obstacles, enabling them to plan and replan their paths efficiently. A comparative analysis between collaborative and standalone perception approaches demonstrates the advantages of this framework in optimizing operational metrics such as task completion time and energy efficiency which in turn increases throughput and reduces costs.

This paper aims to describe the collaborative perception framework, detail the specific features implemented in each case study, and present experimental results demonstrating the effectiveness of the framework. Through these case studies, we highlight the potential of collaborative perception in enhancing multi-robot systems and its implications for future research and practical applications.

\section{Related Work}

\begin{figure*}[!t]
  \centering
  \includegraphics[width=\textwidth]{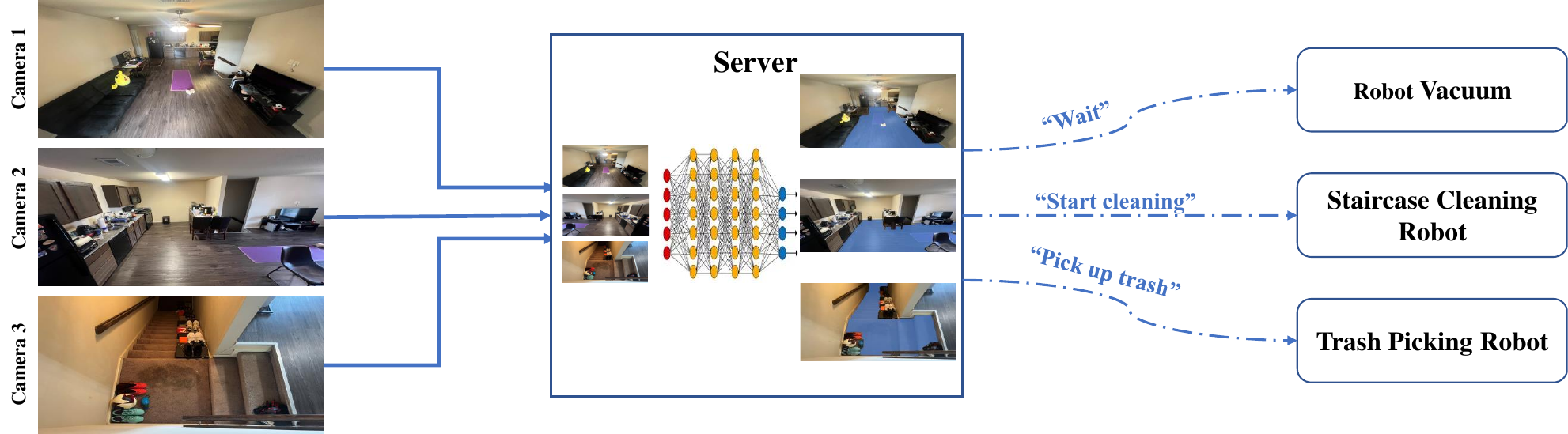}
  \caption{Pipeline for multi-robot decision making based on collaborative perception.}
  \label{fig:cleaning_collab_pipeline}
\end{figure*}

Collaborative perception in multi-robot systems has been explored extensively, focusing on different data fusion techniques \cite{4_liu2023towards}. Early fusion \cite{5_shi2022vips} involves combining raw sensor data from multiple sources at the initial stages of data processing, providing rich environmental representations but requiring high-bandwidth communication. Intermediate fusion \cite{6_wang2020v2vnet, 7_xu2022opv2v} combines partially processed data, balancing data richness and communication efficiency. Late fusion \cite{8_rauch2012car2x, 9_chen2022learning} shares high-level processed information, reducing communication load but potentially losing some detail in environmental representation.

Sharing perception data from multiple sources to a central server provides comprehensive information about the surroundings, facilitating better decision-making for the robots. However, this introduces security concerns, as the data being transmitted and stored centrally can be vulnerable to cyber-attacks. Robust data encryption and secure communication channels are critical to safeguarding the integrity and privacy of sensory data for such safety-critical applications \cite{10_li2023among, 11_tu2021adversarial}.

One of the most widely researched application of collaborative perception is in autonomous driving. In this context, vehicles and roadside infrastructure share sensory data to improve situational awareness and decision-making. Vehicle-to-Vehicle (V2V) \cite{12_ngo2023cooperative}, Vehicle-to-Infrastructure (V2I) \cite{13_mo2024enhanced} and Vehicle-to-Everything (V2X) \cite{14_huang2023v2x} communication systems enable real-time data sharing, enhancing the ability of autonomous vehicles to detect and respond to obstacles, pedestrians, and other vehicles. This collaborative approach significantly enhances safety and efficiency in autonomous driving scenarios.

Other potential applications of collaborative perception in multi-robot systems include automated warehouses, smart home environments, human-robot collaboration, environmental monitoring, and search-and-rescue operations. While most collaborative perception is vision-based, utilizing cameras and LIDARs to gather visual data, it can also be tactile-based. Distributed tactile perception \cite{15_hassan2022tactile, 16_eom2021embedded, 17_ye2022soft}, through sensors embedded in robotic skins, provides a more comprehensive understanding of the environment and physical interactions. This enhanced sensory capability improves the robots' performance in tasks such as manufacturing, assembly, and quality inspection, allowing them to handle delicate and complex operations with greater accuracy and efficiency.

\section{Collaborative Perception Framework}

The collaborative perception framework enables effective multi-robot collaboration through several integrated components. Each robot is equipped with perception modules that consist of sensors like cameras and Lidars. Additionally, the environment the robots operate in could be fitted with sensors as well. All or some of this data could be transmitted to a centralized server via low-latency communication network. 

The centralized server can aggregate data from multiple sources and create a unified and comprehensive understanding of the environment. The server can also utilize the data for optimized task allocation for multi-robot systems. This setup also helps in reducing edge computation on the robots as some non-real-time, compute-intensive tasks could be performed on the server.

In this paper, we specifically analyze how sharing data from sensors embedded in the robot's working environment enhances robot productivity. Figure \ref{fig:cleaning_collab_pipeline}  illustrates a typical pipeline for multi-robot decision making based on collaborative perception.

\section{Case Study 1: Collaborative Home Cleaning Robots}

In modern smart homes, the integration of multiple robots to handle various cleaning tasks can significantly enhance the efficiency and effectiveness of maintaining a clean living space. Collaborative perception, where robots and sensors in the environment share and integrate sensory data, allows such robots to operate more intelligently and cohesively. This case study explores the implementation of a collaborative perception system involving multiple cleaning robots in a home setting, guided by overhead cameras providing real-time RGB images to optimize the cleaning process. This framework provides the basis for working with these robots, although no physical robots were used in the experiments.

\begin{figure}[h]
\centerline{\includegraphics[width=0.5\textwidth]{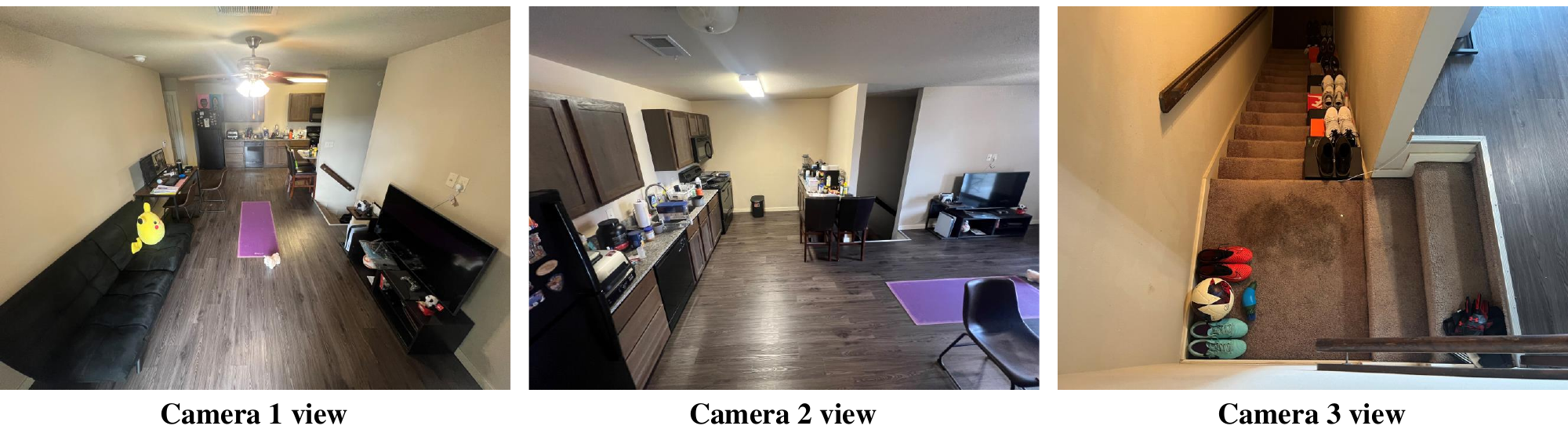}}
\caption{View from overhead cameras}
\label{3_views}
\end{figure}

\subsection{Scenario Description}

A home cleaning scenario is set up with three overhead cameras installed in a section of a living space. Each camera monitors a different section of the space as shown in Fig. \ref{3_views}: one covers one side of the room, another covers the opposite side, and the third monitors the staircase going down to the door. In this scenario, we consider three different cleaning robots: Robotic Vacuum Cleaner (RVC) \cite{irobot_roomba_i5_plus}, Staircase Cleaning Robot (SCR) \cite{sTetro} and Trash Picking Robot (TPR) \cite{bai2018deep}. 
The overhead cameras send data to the server that processes the information to make suitable decisions based on conditions. The room is assumed to be well-lit, as variations in lighting (excessive or insufficient) can significantly impact the performance of deep learning models used for perception \cite{maruschak2024surface}.

\subsection{Collaborative Perception in Home Cleaning}

The collaborative perception system leverages the overhead cameras to instruct specific robots based on the detected needs. The cameras send data in real-time to the server that processes the information. The server interprets the data coming in from the cameras and sends commands to individual robots based on the specific situation. The outline of the pipeline for multi-robot decision making based on collaborative perception is shown in Fig. \ref{fig:cleaning_collab_pipeline}.

\begin{figure}[h]
\centerline{\includegraphics[width=0.5\textwidth]{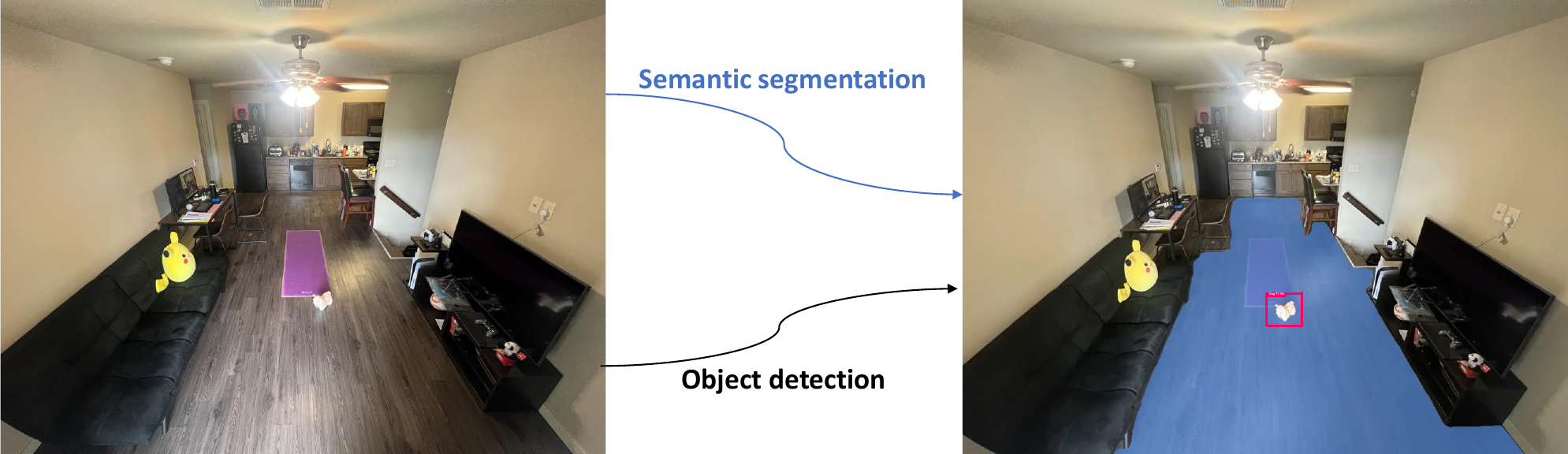}}
\caption{Example output of ML pipeline run on the server}
\label{yolo_demo}
\end{figure}

\begin{figure}[h]
\centerline{\includegraphics[width=0.5\textwidth]{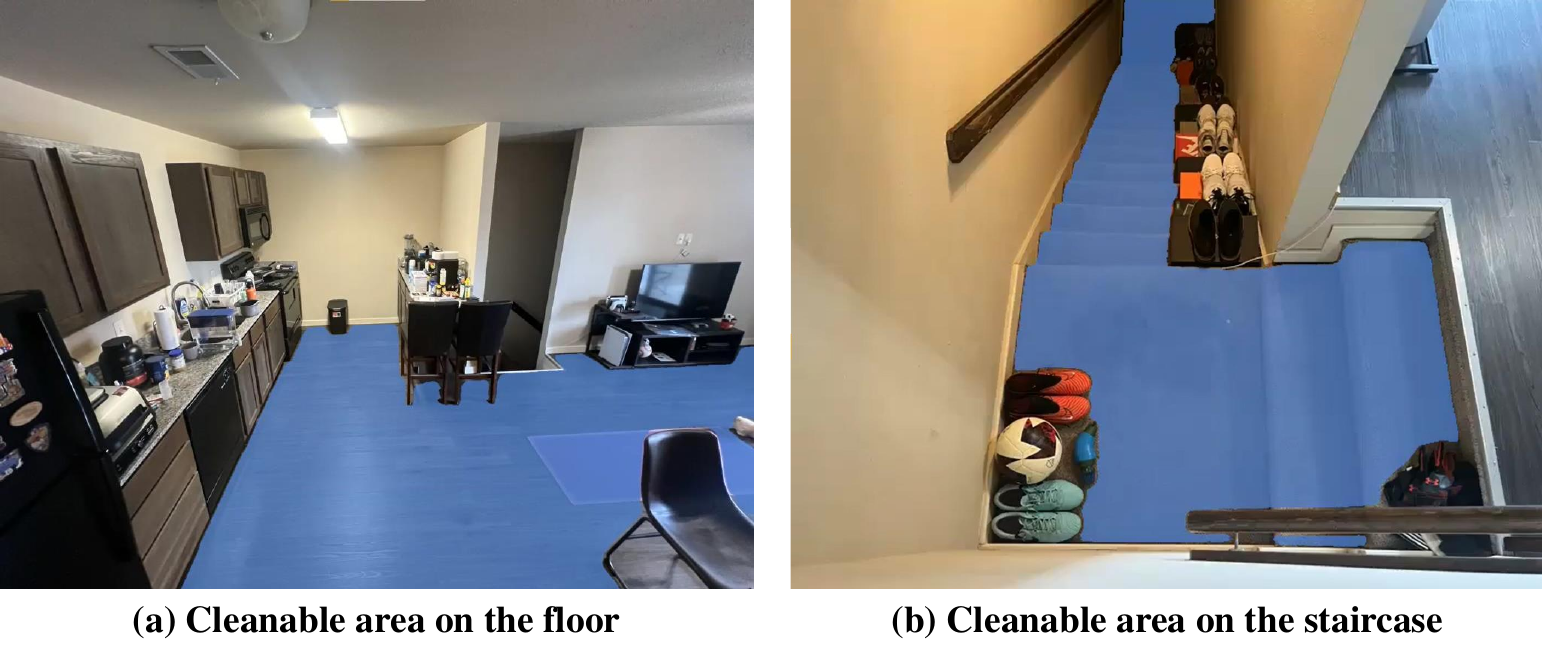}}
\caption{Cleanable area segmentation results}
\label{cleanable_area}
\end{figure}

The server runs a YOLOv8 \cite{yolov8_ultralytics} object detection model to detect the presence of dynamic obstacles like humans and pets. The server also runs a YOLOv8 \cite{yolov8_ultralytics} semantic segmentation model to segment cleanable floor area from the camera feed. Figure \ref{yolo_demo} shows an example of an output of the Machine Learning (ML) pipeline run on the server. The blue mask in Fig. \ref{yolo_demo} represents the cleanable area, while the red square indicates an obstacle that was detected. Figure \ref{cleanable_area} shows the output of the semantic segmentation model accurately segmenting cleanable area for both flat surfaces as well as staircases by avoiding the entities in those areas. The pretrained YOLOv8 models were fine-tuned by training on a dataset of 1000 images. These images were quickly annotated using the Segment Anything Model (SAM) \cite{sam_meta} by Meta. 

\subsection{Conceptual Framework}

Although physical experiments have not yet been conducted, the conceptual framework involves a monitored environment where the room sections are defined and observed by overhead cameras. To ensure safety and cleaning effectiveness, the robots could be programmed to halt operations when humans or pets are detected in the scene. Task prioritization could be implemented so that if trash is detected, the TPR is dispatched first to clear the area. Once the trash is removed, the RVC and SCR proceeds to clean their designated areas, optimizing the cleaning process through coordination.

Potential performance metrics for future experiments include cleaning coverage, cleaning performance, time efficiency, and the effectiveness of the robots' coordination. This framework highlights the potential benefits of collaborative perception in enhancing the efficiency and safety of home cleaning operations through the integrated efforts of multiple robots.

\section{Case Study 2: AMRs in Warehouses}

In warehouse environments, Autonomous Mobile Robots (AMRs) play a crucial role in optimizing logistics and inventory management. This case study explores the implementation of collaborative perception in a warehouse setting to enhance the efficiency and safety of robotic operations.

\subsection{Scenario Description}
A warehouse with multiple racks is simulated. There are AMRs in the warehouse that receive missions to carry pallets or goods autonomously from arbitrary start points to arbitrary goal points depending on specific requirements. For standalone perception, the sensors are only attached to the AMRs. For collaborative perception, there are overhead Lidars, with one installed in every aisle in addition to the sensor suite on the AMRs. This setup allows for real-time sharing of perception data to improve navigation and task execution.

\begin{figure}[h]
\centerline{\includegraphics[width=0.47\textwidth]{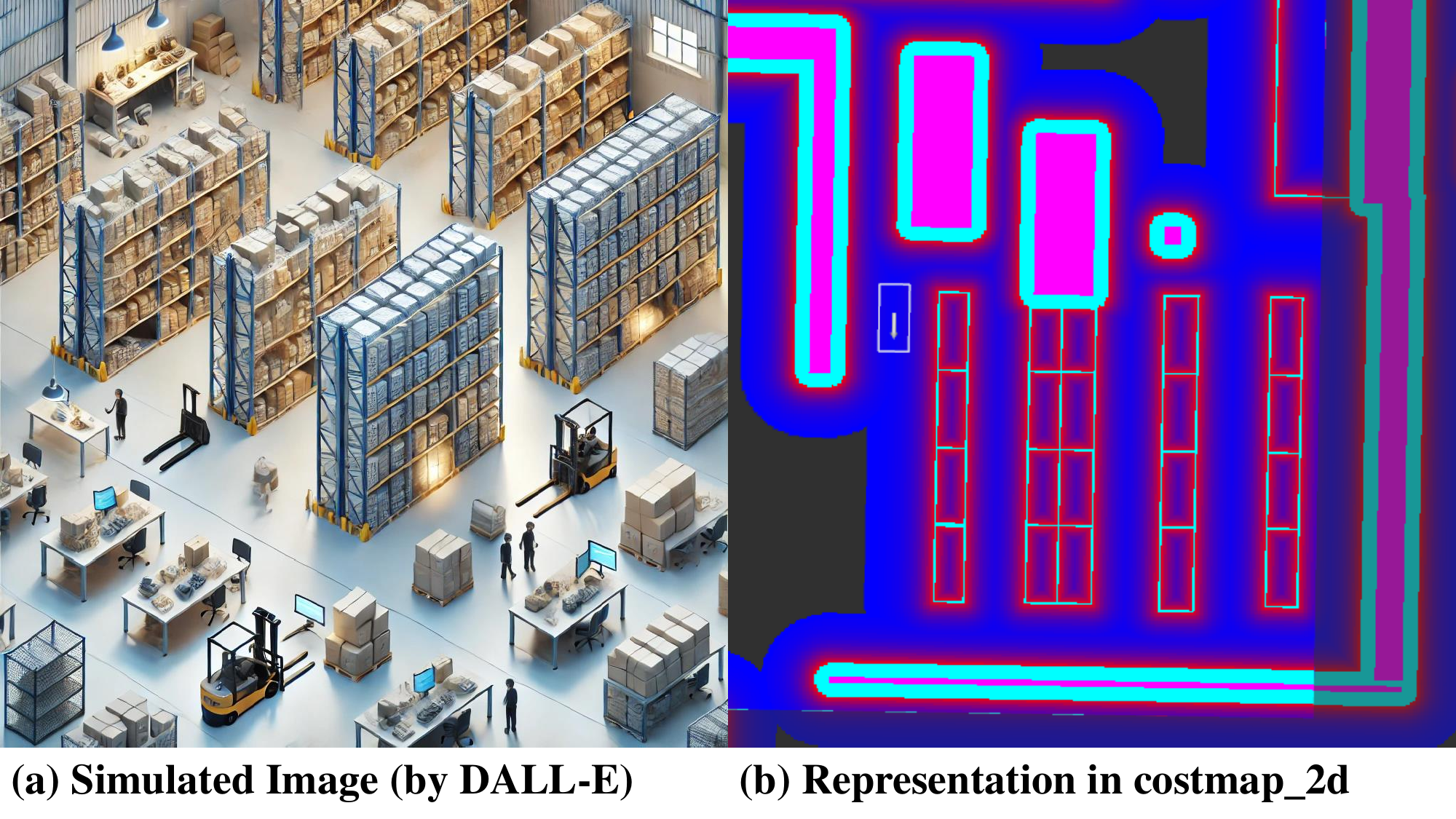}}
\caption{Warehouse environment used for analysis}
\label{real_costmap}
\end{figure}

\subsection{Collaborative Perception in Warehouses}

The robots have a map of the warehouse facility that they use for path planning. The costmap\_2d package \cite{ros_costmap_2d} in Robot Operating System (ROS) is utilized for this purpose. Static obstacles, such as racks and poles, and dynamic obstacles, such as humans, other AMRs, and various entities, are added into the costmap. The picture of the warehouse setting and the corresponding costmap can be seen in Fig. \ref{real_costmap}. In the costmap, black represents free space, dark blue indicates navigable areas close to obstacles, cyan marks the outer boundaries of inflation zones, and bright magenta signifies obstacles. The visualization is done in RViz.

In the standalone perception mode, the costmap is only updated when the individual robot perceives obstacles with its sensors. In this method, only the obstacles that are nearby, within the field of view and that are not occluded are added to the costmap. 

In contrast, for the collaborative perception mode, the individual robots as well as the overhead sensors can add obstacles to the costmap in real-time. The perception data from the overhead sensors are shared to a central server that updates the costmap with any obstacles. The AMRs receive this costmap that contains all the static obstacles and dynamic obstacles seen by the overhead sensors. The AMRs then add the dynamic obstacles that it perceives as well to the local costmap.

Once the AMRs receive a mission to go to a goal point, the AMRs find an optimal path to reach its goal using the ARA\( ^*\)  \cite{likhachev2003ara} algorithm in Search-Based Planning Library \cite{ros_sbpl} of ROS.

\subsection{Experimental Setup and Results}

\begin{figure}[h]
\centerline{\includegraphics[width=0.47\textwidth]{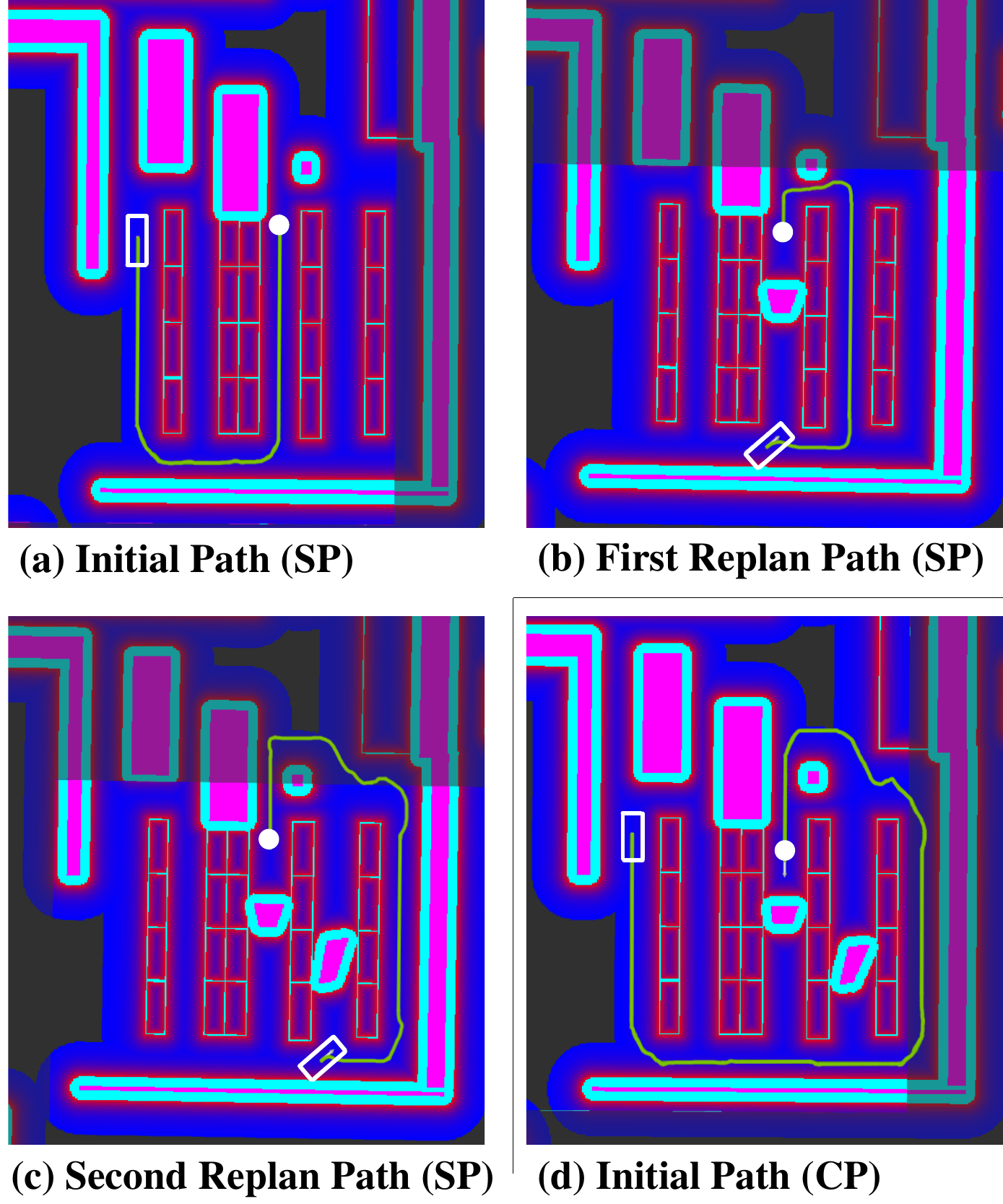}}
\caption{Multiple plans generated to reach goal by Standalone Perception (SP) vs Single plan generated to reach goal by Collaborative Perception (CP)}
\label{SPvsCP}
\end{figure}

50 trials were set up in simulation for a single AMR to navigate from randomly selected feasible start points to randomly selected feasible goal points in the environment shown in Fig. \ref{real_costmap}. Two obstacles were randomly placed in different parts of the map for each trial. Three different metrics were considered for the comparative analysis between standalone perception and collaborative perception framework: Total distance traveled by the robot to reach its goal, Total time taken by the robot to reach its goal and Number of times replanning was necessary.

The results of one of the trials are discussed below and illustrated in Fig. \ref{SPvsCP}. The AMR receives a mission to travel to the goal point indicated by the white circle. In the case of Standalone Perception (SP), the AMR plans a simple path shown in Fig. \ref{SPvsCP}(a). As the AMR approaches the aisle, it encounters an obstacle blocking the path, stops, and replans a new path through the next aisle as shown in Fig. \ref{SPvsCP}(b). The AMR now executes the new plan and encounters another obstacle as it turned into the next aisle, this caused the AMR to stop again and replan a third path to the goal as seen in Fig. \ref{SPvsCP}(c). The AMR ultimately reaches the goal following the latest plan. In contrast, Fig. \ref{SPvsCP}(d) illustrates the Collaborative Perception (CP) scenario, where the overhead Lidars in each aisle updated the costmap with obstacles in real-time. This allows the AMR to plan an uninterrupted path to the goal without the need for replanning. 


\begin{figure}[H]
\centerline{\includegraphics[width=0.47\textwidth]{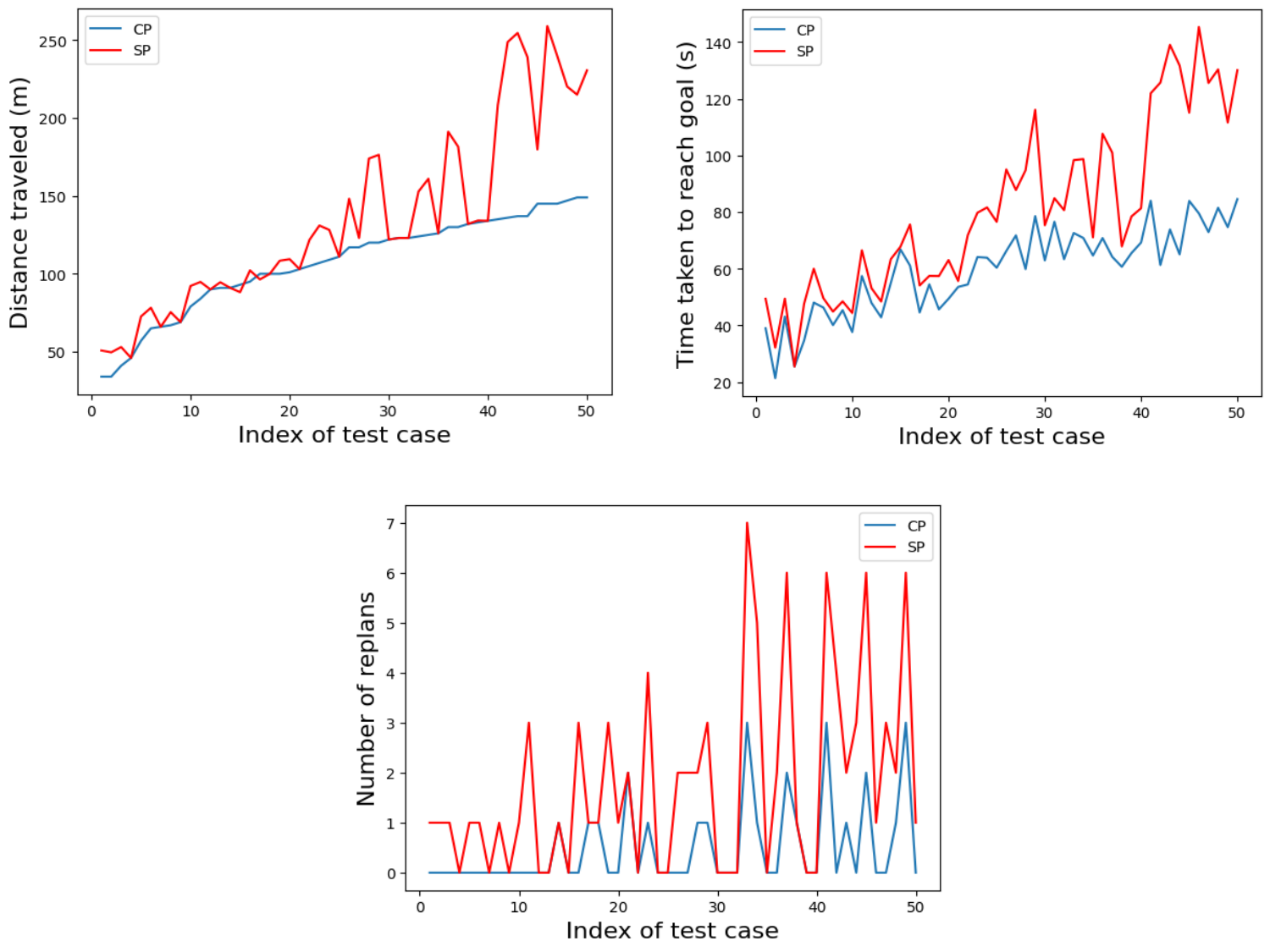}}
\caption{Comparison of different metrics for Standalone Perception (SP) vs Collaborative Perception (CP)}
\label{AMR_metrics}
\end{figure}

Various metrics were tracked during the experiments to compare Collaborative Perception (CP) and Standalone Perception (SP) for AMRs. The distance traveled by AMRs to reach goal locations was recorded across multiple trials. Results show that AMRs consistently travel shorter distances using CP, averaging 37.21 meters less than SP, except when obstacles are not on the shortest path to the goal. Similarly, the average time taken to reach the goal, which includes travel and path planning/replanning time, was reduced with CP by an average of 23.15 seconds.  It was observed that the difference between the distance travelled and time taken to reach the goal using SP and CP increased the farther the goal location was from the start location. Furthermore, the frequency of replanning events was lower with CP, with an average reduction of 1.6 replans, demonstrating its efficiency in navigation. The results of the full set of trials is shown in Fig. \ref{AMR_metrics}. The results also indicate that performance improvements gained from using CP over SP become more pronounced when the AMRs operate in larger warehouse environments.

\section{Discussion}

Using the Collaborative Perception framework provides a more complete information about the robot's working environment. This richer information in turn leads to significant improvements in overall productivity of the team of robots and better task efficiency for individual robots. 

The case study on Collaborative Home Cleaning Robots demonstrates that Collaborative Perception enables efficient task allocation, optimizing cleaning paths and minimizing redundant efforts. This process saves time and enhances safety by detecting humans and pets, ensuring secure operations. Furthermore, the automation of cleaning tasks reduces the need for manual intervention, allowing residents to maintain cleaner living spaces with minimal effort.

The case study on AMRs in warehouses quantitatively proved that leveraging CP by utilizing overhead sensors significantly reduces the distance traveled and time taken by AMRs to complete tasks. This increase in efficiency leads to higher throughput, which more than compensates for the additional cost of the sensors. Moreover, the enhanced operational efficiency boosts overall productivity, resulting in faster processing times and reduced labor costs.

This paper focuses on the sharing of environmental sensor data into a centralized server, which in turn provides actionable insights to individual robots. The framework can be extended to include data sharing between robots, further enhancing the overall system's situational awareness. However, expanding data sharing to this extent introduces challenges such as managing high bandwidth requirements, addressing latency issues, and ensuring data security. Additionally, implementing lossless data compression techniques will be essential to handle the increased data load optimally while maintaining data integrity. Future work will need to address these challenges to fully realize the potential of collaborative perception in multi-robot systems.

The collaborative perception framework has practical applications beyond home cleaning and warehouse automation. In healthcare, it can enhance patient monitoring and robotic assistance in hospitals. In agriculture, it can optimize farming operations through precise monitoring of crops and coordination of agricultural robots. In public safety, it can improve search and rescue missions by providing real-time data and coordination among rescue robots. In surveillance, it can enhance monitoring and security operations by providing comprehensive environmental awareness, enabling quicker respose times, and improving the accuracy of threat detection. Thus, the implementation of collaborative perception can bring substantial benefits, transforming how industries such as healthcare, agriculture, and public safety operate.

\section{Conclusion}

The collaborative perception framework presented in this paper demonstrates significant potential for enhancing multi-robot collaboration through sensor data integration and centralized processing. By aggregating data from various sensors and utilizing advanced algorithms, the framework improves task efficiency, coverage, and safety. Simulation experiments conducted for the AMR case study showed promising results in optimizing robot operations in a warehouse environment. Although physical experiments were not conducted for the home cleaning scenario, the conceptual framework illustrates improvements in cleaning efficiency and potential performance metrics provide a solid foundation for future research. Future work will focus on refining the framework and conducting empirical studies to validate its effectiveness across different applications, addressing challenges such as high bandwidth management, latency issues, and data security. This research underscores the transformative potential of multi-robot systems in enhancing operational efficiency and safety across multiple domains.

\bibliographystyle{IEEEtran}
\bibliography{references}

\end{document}